\documentclass[10pt,conference]{IEEEtran}
\IEEEoverridecommandlockouts
\usepackage{amsmath,amssymb,amsfonts}
\usepackage{graphicx}
\usepackage{booktabs}
\usepackage{array}
\usepackage[table]{xcolor}
\definecolor{bestcell}{RGB}{219,235,255}
\definecolor{secondcell}{RGB}{225,245,225}
\usepackage{url}
\usepackage[numbers,sort&compress]{natbib}
\usepackage{balance}
\usepackage{microtype}
\usepackage[hidelinks]{hyperref}
\newcommand{\figref}[1]{\hyperref[#1]{Figure~\ref*{#1}}}
\newcommand{\tabref}[1]{\hyperref[#1]{Table~\ref*{#1}}}
\newcounter{algorithm}
\renewcommand{\thealgorithm}{\arabic{algorithm}}
\title{Structural Conditional Independence Test\\via Neural Quantile Resolution}
\author{%
\IEEEauthorblockN{Khoa Nguyen, Bao Duong, Thin Nguyen}
\IEEEauthorblockA{Deakin Applied Artificial Intelligence Initiative, Geelong, Australia\\
\{khoa.nguyen, b.duong, thin.nguyen\}@deakin.edu.au}
}
\begin{document}
\maketitle

\begin{abstract}
Conditional independence testing is difficult when dependence is weak, localized, or expressed through distributional changes rather than a global mean shift. Existing tests usually return a scalar statistic or p-value, which can detect rejection but gives limited resolution about where the conditional law changes. We propose QRCIT, a quantile-resolution conditional independence test that compares \(P(X\mid Z)\) and \(P(X\mid Y,Z)\) through an amortized quantile deformation. A lightweight residual quantile-flow model first learns the baseline conditional quantile law and then learns the additional deformation induced by \(Y\), yielding both a calibrated global test statistic and an inspectable deformation field over context and quantile level. We also introduce the Structural Dependency Benchmark (SDB), which evaluates real-world-inspired dependence structures, such as context-local, tail-only, sign-reversing, moving-ridge, and mean-neutral patterns. Across synthetic PNL and SDB benchmarks, QRCIT improves detection of structured alternatives while maintaining empirical Type-I control. Increasing quantile resolution consistently improves merged-SDB power, and the local-rescue strategy sharpens detection when dependence is concentrated in structural regions. On Sachs, METABRIC breast cancer, and CCLE/PLX4720, QRCIT provides competitive real-data performance and interpretable deformation patterns. These results support quantile-resolved conditional-law deformation as a practical framework for testing and diagnosing structured conditional dependence.
Code is available at \href{https://github.com/khoangdadk/qrcit}{\textcolor{blue}{\nolinkurl{https://github.com/khoangdadk/qrcit}}}.
\end{abstract}

\section{Introduction}

Conditional independence testing (CIT) asks whether
\begin{equation}
H_0: X \perp\!\!\!\perp Y \mid Z
\end{equation}
should be rejected, and is a basic primitive in causal discovery~\cite{zhang2012kci}, feature screening and controlled variable selection~\cite{candes2018knockoffs}, biological dependency analysis~\cite{giorgi2014cindy,bellot2019gcit}, fairness auditing~\cite{hardt2016equality}, and robust or invariant prediction~\cite{peters2016invariant}. Existing CIT procedures usually compress the sample into a scalar statistic and a p-value. This is necessary for hypothesis testing, but it hides structure: a weak statistic may reflect true conditional independence, low power against a localized alternative, cancellation between opposite effects, tail-only dependence, or a distributional shape change with little conditional-mean shift. Conversely, a rejection rarely explains where or how the conditional law changed.

\begin{figure*}[t]
\centering
\includegraphics[width=\textwidth]{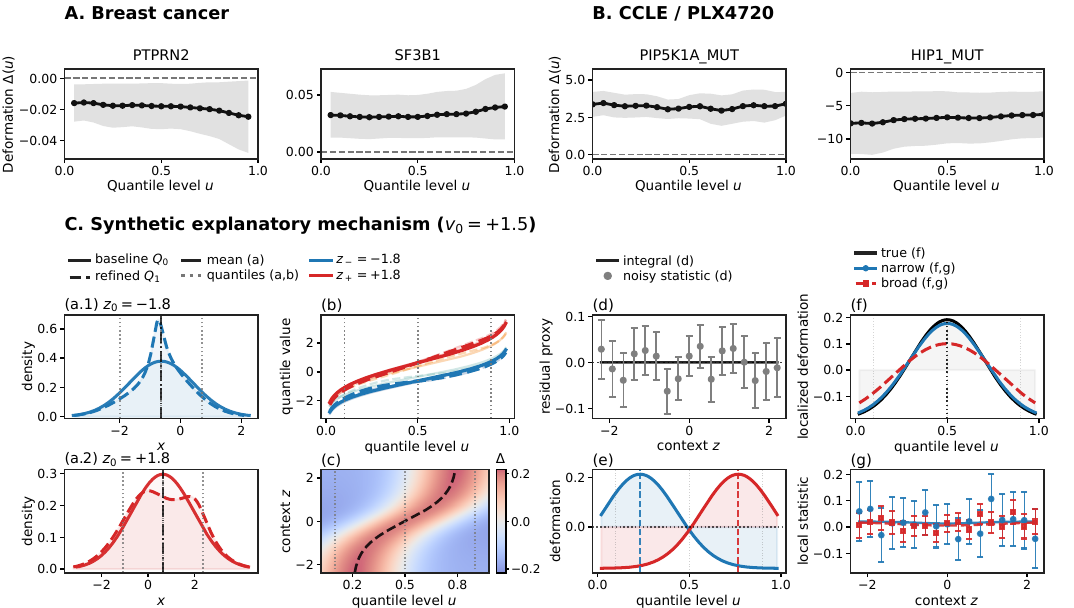}
\caption{Real-data evidence and synthetic explanation for quantile-resolved conditional dependence. In A--B, black curves show mean conditional quantile contrasts, averaged over held-out contexts and across folds; grey bands show \(\pm1\) standard deviation across folds. A contrasts gene expression at the 75th versus 25th percentile; B contrasts mutation presence versus absence. Each contrast compares the learned quantiles at two values of \(Y\), so the shared baseline cancels. Positive and negative values indicate upward and downward quantile shifts, respectively. (A) Breast-cancer deformation profiles in the cBioPortal-reconstructed METABRIC benchmark. PTPRN2 shows negative upper-tail tumor-size deformation, interpreted as a progression-associated conditional signature~\cite{tavazoie2008micrornas}; SF3B1 shows positive upper-tail deformation, consistent with breast-cancer relevance of splicing-factor dysregulation~\cite{tcga2012breast}. (B) CCLE/PLX4720 mutation-response profiles. PIP5K1A\_MUT shows a broadly positive fitted Amax-scale contrast, while HIP1\_MUT shows a negative contrast whose magnitude varies across quantile levels. These profiles describe how mutation status shifts the fitted response quantiles conditional on the other selected mutations. (C) A synthetic mean-neutral example in which observing \(v_0\), playing the role of \(Y\), changes the conditional distribution's shape with little change in its mean. Panels (a.1)--(a.2) show baseline and refined density slices; (b) shows the corresponding quantile functions; and (c) shows the signed deformation \(\Delta(v_0,z,u)=Q_1(v_0,z,u)-Q_0(z,u)\). Panels (d)--(g) illustrate the difficulty of detecting these changes: mean-like summaries can be weak, broad neighborhoods blur local structure, and narrow neighborhoods increase estimation variance~\cite{he2025hardness}.}
\label{fig:intro}
\end{figure*}

\figref{fig:intro} illustrates this motivation. Panels A--B show the direction and quantile-specific shape of fitted real-data conditional-law contrasts. Panel C gives a controlled mean-neutral structural example: density and quantile views reveal changes in the conditional distribution, while residual-covariance and local scalar summaries can be weak or unstable. Together, these panels illustrate the central gap addressed by QRCIT: conditional dependence is not only a question of whether \(X\perp\!\!\!\perp Y\mid Z\) is rejected, but also where the conditional-law discrepancy appears over context and quantile level. This motivates a test that keeps a calibrated global p-value while retaining a deformation field for inspection, consistent with local and context-specific independence settings where dependence can be restricted to part of the conditioning space~\cite{kim2021localperm,hwang2023local}.

Such structural discrepancies arise across data-mining applications. In genomics and pharmacogenomics, regulatory or drug-response signals can depend on molecular context, corresponding to context-local dependence~\cite{giorgi2014cindy,barretina2012ccle}. In finance, systemic risk and growth-at-risk are explicitly tail-focused, motivating tests that do not average away lower-tail dependence~\cite{adrian2016covar,adrian2019vulnerable}. In policy and social-science applications, treatment effects may vary across the outcome distribution, so mean-level summaries can miss distributional or sign-varying effects~\cite{chernozhukov2005ivqr,bitler2016headstart}. These examples require a CIT method to expose how \(P(X\mid Y,Z)\) differs from \(P(X\mid Z)\), not only whether the difference survives scalar aggregation.

The theoretical difficulty is that CIT compares conditional objects across \(Z\). With continuous or high-dimensional \(Z\), exact local replication is unavailable, so tests approximate locality through smoothing, kernels, regression, embeddings, or learned conditional distributions. This creates a stability--resolution tradeoff: broad neighborhoods reduce variance but blur localized dependence, whereas narrow neighborhoods preserve structure but may contain too few observations. Since conditional-object estimation and conditioning-kernel choices affect both Type-I error and power~\cite{he2025hardness}, practical CIT should retain a resolution layer that distinguishes true independence from localized, tail-specific, or cancelling dependence.

This paper asks whether CIT can be equipped with such a resolution layer. Our starting point is the law-level characterization
\begin{equation}
X \perp\!\!\!\perp Y \mid Z
    \quad \Longleftrightarrow \quad
    P_{X\mid Y,Z}=P_{X\mid Z}\quad\text{a.s.}
\end{equation}
For scalar \(X\), let
\begin{equation}
\begin{aligned}
Q_0(z,u) &= Q_{X\mid Z=z}(u),\\
Q_1(y,z,u) &= Q_{X\mid Y=y,Z=z}(u),
\end{aligned}
\end{equation}
where \(u\in(0,1)\) is the quantile level. We define the quantile-flow deformation
\begin{equation}
\Delta(y,z,u)=Q_1(y,z,u)-Q_0(z,u).
\end{equation}
Under conditional independence, this field is zero almost everywhere; under conditional dependence, its geometry describes how observing \(Y\) deforms the conditional law of \(X\) beyond \(Z\).

Inspired by residual networks and residual flows~\cite{behrmann2019invertible,chen2019residual}, we propose \textbf{Quantile-Resolution Conditional Independence Test (QRCIT)}, a residual neural quantile-flow framework. QRCIT learns a baseline conditional quantile model \(\widehat Q_0(Z,u)\) and only the additional deformation \(D_\eta(Y,Z,u)\), giving
\begin{equation}
\widehat Q_1(Y,Z,u)=\widehat Q_0(Z,u)+D_\eta(Y,Z,u).
\end{equation}
The global statistic measures predictive gain under the pinball loss: whether adding \(Y\) improves quantile prediction of \(X\) beyond \(Z\). Its p-value is calibrated by conditional randomization, while the learned deformation field remains available for structural inspection. Thus QRCIT keeps the standard hypothesis-testing interface but exposes the conditional-law discrepancy before it is aggregated into a scalar statistic.

The central design principle is \emph{resolution}. Prior Wasserstein CIT uses multiresolution distributional comparisons within conditioning-space bins~\cite{neykov2024wasserstein}. QRCIT adopts the broad resolution principle along a different axis: a quantile grid \(\mathcal U_K=\{u_1,\ldots,u_K\}\) evaluates conditional-law change. A coarse grid summarizes broad discrepancy, while a finer grid can resolve localized changes. Since \(u\) is an input to the neural model, increasing \(K\) refines evaluation resolution without increasing model parameters.

Our contributions are as follows.
\begin{itemize}
    \item We represent conditional-law discrepancy through an inspectable quantile-deformation field, building on the quantile characterization of conditional independence.
    \item We propose \textbf{QRCIT}, a residual neural quantile-flow test that learns \(\widehat Q_0(Z,u)\) and \(D_\eta(Y,Z,u)\), tests via predictive gain, and calibrates global p-values by conditional randomization.
    \item We identify quantile resolution as a practical degree of freedom for CIT: the grid size controls how finely QRCIT inspects localized distributional alternatives without changing the model size.
    \item We introduce \textbf{SDB}, a controlled benchmark suite for structured conditional-law changes motivated by practical domain evidence and the stability--resolution difficulty of conditional testing.
    \item We evaluate QRCIT on synthetic, SDB, sliced multivariate, and biological benchmarks, showing improved detection of structured alternatives and interpretable real-data deformation patterns.
\end{itemize}

\section{Background and Related Work}
\label{sec:related}

We position QRCIT relative to classical and quantile-based CIT, neural conditional-randomization and flow-based tests, and hardness results motivating structural resolution.

\subsection{Conditional Independence Testing}

Conditional independence testing has been developed through kernel, distance, and regression-based formulations. Kernel-based conditional independence (KCI) tests measure conditional dependence with RKHS statistics and are widely used in causal discovery~\cite{zhang2012kci}. Nearly minimax Wasserstein CIT uses weighted multiresolution distributional comparisons within conditioning-space bins~\cite{neykov2024wasserstein}. Regression-residual methods provide a complementary route: the generalized covariance measure (GCM) regresses \(X\) and \(Y\) on \(Z\) and tests residual covariance, with validity depending mainly on conditional-mean estimation~\cite{shah2020gcm}. Representation-learning approaches such as LCIT learn latent representations of \(X\) and \(Y\) that remove information about \(Z\), and then test whether the remaining latent variables are dependent~\cite{duong2022lcit}.

Quantile-based CIT has established foundations. Su and White~\cite{su2012quantile} compare local-polynomial conditional quantile estimates, while Petersen and Hansen~\cite{petersen2021partial} use quantile regression to construct partial-copula residuals. Quantile graphical models characterize conditional dependence across quantile levels~\cite{belloni2025qgm}. QRCIT's emphasis is a shared residual quantile-deformation model that supports predictive-gain testing while retaining a signed field for diagnosis and adjustable quantile evaluation resolution.

\subsection{Neural Conditional Randomization and Flow-Based Tests}

The conditional randomization test (CRT) calibrates a statistic by resampling one variable from an estimated null conditional distribution. Model-powered CIT and GCIT instantiate this principle with predictive or generative models; GCIT uses a generative adversarial network to approximate the null distribution and evaluate test statistics against generated null samples~\cite{sen2017mpcit,bellot2019gcit}. More recent neural CIT methods replace GANs with conditional diffusion or flow models. CDCIT uses conditional diffusion models for CRT-style testing with a classifier-based conditional mutual information statistic~\cite{yang2024cdcit}. Transport-map CIT estimates conditional continuous normalizing flows to transform conditional independence into unconditional independence, while FMCIT uses flow matching to accelerate repeated CI queries in causal discovery~\cite{he2025transport,zhao2026fmcit}. These methods improve testing by learning conditional distributions or transports. QRCIT differs in its learned object: it does not aim to learn a full conditional density or global transport map, but a scalar-valued quantile deformation of \(X\) that remains inspectable after computing the global p-value.

\subsection{Hardness and Structural Resolution}

Hardness results motivate explicit assumptions and target alternatives in CIT. Shah and Peters show that, without restrictions, no universally valid finite-sample conditional independence test can have nontrivial power, and propose GCM as a residual test under weaker conditional-mean estimation requirements~\cite{shah2020gcm}. Recent practical analysis further shows that KCI Type-I behavior is strongly affected by conditional mean embedding estimation error, while power depends critically on the conditioning kernel; choices that improve power can also inflate Type-I error~\cite{he2025hardness}. These results expose a stability--resolution tension: conditioning must be local enough to reveal dependence, but stable enough to avoid spurious rejection. This view is consistent with local permutation tests, which compare observations with nearby conditioning values, and recent continuous-variable local-independence work, which seeks context regions where independence relations change~\cite{kim2021localperm,hwang2023local}. QRCIT differs by keeping a global CRT-calibrated p-value while exposing a quantile--context deformation field for regional inspection.

The structural alternatives in SDB instantiate concrete versions of this hardness mechanism. A context-local alternative can be expressed as a quantile deformation
\begin{equation}
\Delta(y,z,u)=a(y,u)\mathbf 1\{z\in A\},
    \qquad \mathbb P(Z\in A)=\pi ,
\end{equation}
so that dependence is confined to a small context region \(A\). Broad conditioning neighborhoods mix \(A\) with its complement and dilute the discrepancy, whereas narrow neighborhoods reduce the effective sample size to the order of \(n h^{d_Z}\), exposing the bias--variance tradeoff of localized CIT. Tail-only alternatives create an analogous resolution problem along the response distribution: if
\begin{equation}
\Delta(y,z,u)=a(y,z)\mathbf 1\{u>\tau\},
    \qquad \tau\approx 0.8 \text{ or } 0.9,
\end{equation}
then a statistic aggregating across quantile levels observes only a fraction \(1-\tau\) of the deformation, while a local quantile-context estimator effectively operates with sample size \(n h^{d_Z}(1-\tau)\). These examples illustrate why SDB varies not only signal strength but also the location of conditional-law discrepancies over \(Z\) and \(u\). QRCIT addresses this structured regime by estimating an amortized deformation \(\Delta_\theta(y,z,u)\), retaining resolution over context and quantile level while pooling information through a shared model.

\section{Quantile-Resolution Conditional Independence Test}
\label{sec:method}

\smallskip\noindent\textbf{Overview.}
QRCIT tests conditional independence while retaining a resolution layer for the conditional-law discrepancy. The method has four components. First, it represents conditional dependence as a quantile-flow deformation between \(P(X\mid Z)\) and \(P(X\mid Y,Z)\). Second, it learns this deformation with a residual quantile-flow model. Third, it aggregates quantile-level predictive gain into a global statistic calibrated by conditional randomization. Fourth, it uses the quantile grid and local regions to inspect where the conditional-law change is concentrated. Algorithm~\ref{alg:qrcit} summarizes the training, calibration, and output steps. We present the scalar-\(X\) case first and describe a sliced multivariate extension at the end.

\smallskip\noindent\textbf{Problem setup.}
Let \(\mathcal D=\{(X_i,Y_i,Z_i)\}_{i=1}^n\) be i.i.d. samples, where \(X\) is initially scalar, \(Y\) is the candidate variable, and \(Z\) is the conditioning context. The null hypothesis is
\begin{equation}
H_0: X\perp\!\!\!\perp Y\mid Z.
\end{equation}
QRCIT keeps the usual CIT goal of testing this null, but it also estimates how the conditional law of \(X\) changes when \(Y\) is revealed beyond \(Z\).

\smallskip\noindent\textbf{Quantile-flow deformation.}
For \(u\in(0,1)\), let
\begin{equation}
Q_0(z,u)=Q_{X\mid Z=z}(u),
    \qquad
    Q_1(y,z,u)=Q_{X\mid Y=y,Z=z}(u),
\end{equation}
where quantiles are understood as generalized inverse CDFs. Define
\begin{equation}
\Delta(y,z,u)=Q_1(y,z,u)-Q_0(z,u).
\end{equation}
The next result restates the conditional-quantile characterization of independence~\cite{belloni2025qgm} in deformation coordinates.

\noindent\textbf{Proposition 1 (Deformation characterization).}
Assume regular conditional laws \(P_{X\mid Z=z}\) and \(P_{X\mid Y=y,Z=z}\) exist. Then
\begin{equation}
X\perp\!\!\!\perp Y\mid Z
    \quad\Longleftrightarrow\quad
    Q_1(Y,Z,u)=Q_0(Z,u)
\end{equation}
for \(P_{Y,Z}\otimes\lambda\)-almost every \((Y,Z,u)\), where \(\lambda\) is Lebesgue measure on \((0,1)\). Equivalently, \(\Delta(Y,Z,u)=0\) a.e.

\noindent\emph{Sketch.}
Conditional independence is equivalent to \(P_{X\mid Y,Z}=P_{X\mid Z}\) a.s. Equality of one-dimensional distributions is equivalent to equality of generalized quantile functions for almost every \(u\). Subtracting the two quantile functions gives the result.

\smallskip\noindent\textbf{Residual quantile-flow model.}
Proposition~1 motivates a residual design in which the conditional law after adding \(Y\) is expressed as a deformation of the baseline law. We parameterize
\begin{equation}
\widehat Q_1(Y,Z,u)=\widehat Q_0(Z,u)+D_\eta(Y,Z,u),
\end{equation}
where \(D_\eta\) is the learnable counterpart of \(\Delta\). The additive residual form is inspired by residual parameterizations in invertible and flow-based generative models, where a learned residual component provides a flexible but stable deformation of a baseline transformation~\citep{behrmann2019invertible,chen2019residual}. In QRCIT, this design serves the same purpose at the level of conditional quantiles: \(\widehat Q_0(Z,u)\) represents the baseline conditional quantile law, while \(D_\eta(Y,Z,u)\) efficiently learns the additional deformation associated with adding \(Y\). Under \(H_0\), no residual correction should be needed; under the alternative, \(D_\eta\) learns where and how \(Y\) changes the conditional law of \(X\) beyond \(Z\).

Let
\begin{equation}
\rho_u(r)=r\{u-\mathbf 1(r<0)\}
\end{equation}
be the pinball loss. QRCIT trains \(\widehat Q_0\) and \(D_\eta\) by minimizing
\begin{equation}
\begin{aligned}
    \mathcal L
    =
    \mathbb E_{(X,Y,Z),u}
    \big[
        &\rho_u\{X-\widehat Q_0(Z,u)\}
        +\!
        \rho_u\{X-\widehat Q_1(Y,Z,u)\}
    \big]  \\
    &+
    \lambda_{\mathrm{mono}}\mathcal R_{\mathrm{mono}}
    +
    \lambda_D\mathcal R_D .
\end{aligned}
\end{equation}
The two regularizers have different roles. Let \([a]_+=\max(a,0)\). The monotonicity penalty is the continuous crossing penalty
\begin{equation}
\begin{aligned}
\mathcal R_{\mathrm{mono}}
&=
\mathbb E_Z\!\int_0^1
\!\left[-\partial_u \widehat Q_0(Z,u)\right]_+^2du\\
&\quad+
\mathbb E_{Y,Z}\!\int_0^1
\!\left[-\partial_u \widehat Q_1(Y,Z,u)\right]_+^2du .
\end{aligned}
\end{equation}
It penalizes negative quantile slopes, encouraging nondecreasing baseline and refined quantile functions. Penalizing crossing is necessary because nonmonotone quantile curves can create artificial deformation patterns in \(\widehat Q_1-\widehat Q_0\). The deformation penalty is
\begin{equation}
\mathcal R_D
=
\mathbb E_{Y,Z}\!\int_0^1
D_\eta(Y,Z,u)^2\,du ,
\end{equation}
which shrinks residual deformation toward the null preference \(D_\eta=0\). This reduces spurious conditional-law change when \(Y\) does not add information beyond \(Z\), while still allowing nonzero deformation when the pinball loss supports it. In the residual-deformation implementation used in our experiments, this objective is optimized in two stages: we first fit \(\widehat Q_0\) using the baseline pinball loss; then we freeze \(\widehat Q_0\) and train the deformation head \(D_\eta\) through \(\widehat Q_1=\widehat Q_0+D_\eta\). This staged optimization matches the statistical decomposition above: the baseline law is estimated first, and the residual module learns only the additional conditional-law change attributable to \(Y\) beyond \(Z\). The full QRCIT workflow is summarized in Algorithm~\ref{alg:qrcit}.

\smallskip\noindent\textbf{Predictive-gain statistic.}
The predictive-gain statistic is meaningful because conditional quantiles minimize pinball risk. Given a grid \(\mathcal U_K=\{u_1,\ldots,u_K\}\), QRCIT measures the gain from adding \(Y\):
\begin{equation}
\begin{aligned}
T_K
&=
\frac{1}{nK}\sum_{i=1}^n\sum_{k=1}^K
\Big[
\rho_{u_k}\{X_i-\widehat Q_0(Z_i,u_k)\}\\
&\hspace{5.6em}
-
\rho_{u_k}\{X_i-\widehat Q_1(Y_i,Z_i,u_k)\}
\Big].
\end{aligned}
\end{equation}
Large \(T_K\) indicates that the refined model using \(Y\) reduces quantile risk relative to the baseline model using only \(Z\).

\noindent\textbf{Lemma 1 (Population predictive-gain target).}
Assume \(\mathbb E|X|<\infty\). For fixed \(u\in(0,1)\), let \(Q_0(Z,u)\) and \(Q_1(Y,Z,u)\) be conditional \(u\)-quantiles of \(X\mid Z\) and \(X\mid Y,Z\). Define
\begin{equation}
g(u)
    =
    \mathbb E\!\left[
    \rho_u\{X-Q_0(Z,u)\}
    -
    \rho_u\{X-Q_1(Y,Z,u)\}
    \right].
\end{equation}
Then \(g(u)\ge 0\). If the conditional \(u\)-quantile is unique, then \(g(u)=0\) iff \(Q_1(Y,Z,u)=Q_0(Z,u)\) a.s. If this equality holds for almost every \(u\), Proposition~1 implies conditional independence.

\noindent\emph{Sketch.}
The pinball loss is minimized by the conditional \(u\)-quantile. Since \((Y,Z)\) is at least as informative as \(Z\), its optimal pinball risk is no larger. Equality under uniqueness implies equality of the two conditional quantiles at that \(u\).

Lemma~1 explains why QRCIT can detect dependence whose signed deformations cancel in a mean summary. At the population level, an upward shift at one quantile and a downward shift at another can each yield a positive reduction in pinball risk. QRCIT aggregates these predictive gains across quantiles, so both shifts contribute evidence even when their average signed deformation is close to zero. The deformation field retains their directions and locations for interpretation.

\smallskip\noindent\textbf{Conditional randomization calibration.}
Calibration is handled by conditional randomization; the oracle case gives the relevant Type-I benchmark. For \(b=1,\ldots,B\), draw
\begin{equation}
\widetilde Y_i^{(b)}\sim \widehat P(Y\mid Z_i),
\end{equation}
compute \(T_K^{(b)}\), and report
\begin{equation}
p
    =
    \frac{
    1+\sum_{b=1}^B \mathbf 1\{T_K^{(b)}\ge T_K^{\mathrm{obs}}\}
    }{B+1}.
\end{equation}

\noindent\textbf{Proposition 2 (Oracle Type-I control).}
Consider an evaluation sample independent of the data used to fit \(\widehat Q_0\), \(D_\eta\), and any nuisance components in the statistic. Suppose null samples are generated from the oracle conditional law \(\widetilde Y_i^{(b)}\sim P(Y\mid Z_i)\). Under \(H_0\), conditional on the fitted models and evaluation \((X_i,Z_i)\), \(T_K^{\mathrm{obs}}\) and \(\{T_K^{(b)}\}_{b=1}^B\) are exchangeable. Hence
\begin{equation}
\Pr_{H_0}(p\le \alpha)\le \alpha.
\end{equation}

\noindent\emph{Sketch.}
Under \(H_0\), \(Y_i\mid Z_i\) has the same law as each oracle draw and is independent of \(X_i\) given \(Z_i\). With fitted models fixed, the statistic is a deterministic function of \((X,Z,Y)\), so the observed and null-resampled statistics are exchangeable. In practice, \(P(Y\mid Z)\) is estimated; calibration then depends on the quality of \(\widehat P(Y\mid Z)\), and empirical size is reported in experiments. Proposition~2 concerns an independent evaluation sample; calibration of the pooled cross-fitting implementation is assessed empirically.

\begin{figure}[t]
\refstepcounter{algorithm}\label{alg:qrcit}
\centering
\begin{minipage}{0.98\linewidth}
\footnotesize
\hrule
\vspace{0.35ex}
\noindent\textbf{Algorithm~\thealgorithm\ QRCIT} for conditional independence testing
\vspace{0.35ex}
\hrule
\vspace{0.45ex}
\noindent\textbf{Input:} data \(\mathcal D=\{(X_i,Y_i,Z_i)\}_{i=1}^n\), quantile grid \(\mathcal U_K\), folds \(F\), null draws \(B\), conditional sampler \(\widehat P(Y\mid Z)\).\\
\textbf{Output:} global p-value \(p_{\mathrm{global}}\), deformation summaries, and optional local-rescue p-values.\\[0.35ex]
\begin{tabular}{@{}r@{\hspace{0.45em}}p{0.86\linewidth}@{}}
1. & Split \(\mathcal D\) into \(F\) folds.\\
2. & \textbf{for} each fold \(f=1,\ldots,F\) \textbf{do}\\
3. & \quad Fit the baseline quantile model \(\widehat Q_0(Z,u)\) on the training folds using pinball loss.\\
4. & \quad Freeze \(\widehat Q_0\) and fit \(D_\eta(Y,Z,u)\) through \(\widehat Q_1=\widehat Q_0+D_\eta\).\\
5. & \quad Compute the held-out statistic \(T_{K,f}^{\mathrm{obs}}\) using the predictive gain over \(\mathcal U_K\).\\
6. & \quad \textbf{for} \(b=1,\ldots,B\) \textbf{do}\\
7. & \quad\quad Draw \(\widetilde Y_i^{(b)}\sim \widehat P(Y_i\mid Z_i)\) for held-out samples.\\
8. & \quad\quad Compute the null statistic \(T_{K,f}^{(b)}\) with \(\widetilde Y^{(b)}\).\\
9. & \quad \textbf{end for}\\
10. & \textbf{end for}\\
11. & Aggregate observed and null statistics across folds and return the CRT p-value.\\
12. & If regions are specified, compute region p-values and deformation summaries for local-rescue inspection.\\
\end{tabular}
\vspace{0.35ex}
\hrule
\end{minipage}
\vspace{-0.5ex}
\end{figure}

\smallskip\noindent\textbf{Quantile resolution, region inspection, and local-rescue test.}
Quantile resolution affects how well localized gain is approximated. The grid size \(K\) controls the resolution at which QRCIT evaluates conditional-law change. At the population level, the global statistic approximates an integral of the gain density \(g(u)\) over the quantile axis.

\noindent\textbf{Proposition 3 (Quantile-grid approximation).}
Let \(g:[a,b]\to\mathbb R\), \(0<a<b<1\), be a bounded-variation population gain density. For an equally spaced grid \(\mathcal U_K=\{u_1,\ldots,u_K\}\subset[a,b]\),
\begin{equation}
\left|
    \frac{b-a}{K}\sum_{k=1}^K g(u_k)
    -
    \int_a^b g(u)\,du
    \right|
    \le
    C\frac{V(g)}{K},
\end{equation}
where \(V(g)\) is the total variation of \(g\) and \(C\) is a universal constant.

\noindent\emph{Sketch.}
This is the standard Riemann-sum approximation bound for bounded-variation functions. If the gain is concentrated on a quantile interval of width \(w\), a grid spacing much larger than \(w\) can under-resolve the active region, while \(K=\Omega((b-a)/w)\) resolves it. The result justifies resolution as an approximation principle, while finite-sample power can still depend on estimation and calibration error. The role of \(K\) is analogous in spirit to multiresolution Wasserstein CIT, where aggregating localized discrepancies across resolutions improves sensitivity to structured alternatives~\cite{neykov2024wasserstein}. In QRCIT, \(K\) controls resolution along the quantile axis: a coarse grid captures broad distributional change, while a finer grid can resolve tail-local or moving-\(u\) deformation.

For a region \(\mathcal R\subseteq\{1,\ldots,n\}\times\mathcal U_K\), QRCIT computes
\begin{equation}
\begin{aligned}
T_{\mathcal R}
&=
\frac{1}{|\mathcal R|}
\sum_{(i,k)\in\mathcal R}
\Big[
\rho_{u_k}\{X_i-\widehat Q_0(Z_i,u_k)\}\\
&\hspace{5.6em}
-
\rho_{u_k}\{X_i-\widehat Q_1(Y_i,Z_i,u_k)\}
\Big].
\end{aligned}
\end{equation}
We also use a deformation-energy regional statistic
\begin{equation}
T_{\mathcal R}^{DE}
=
\frac{1}{|\mathcal R|}
\sum_{(i,k)\in\mathcal R} D_\eta(Y_i,Z_i,u_k)^2 .
\end{equation}
Here \(\mathcal R\) may be context-local (a \(Z_0\) slab), quantile-local (tail \(u_k\)), or both.
Regional aggregation focuses the statistic on observations and quantile levels where the conditional law may change. For example, a tail effect confined to one context subgroup contributes to only a small fraction of the terms in the global average. Restricting the statistic to that subgroup and tail can make the signal more prominent, although a smaller region also leaves fewer observations for estimation. Predictive gain measures whether the added variable improves quantile prediction within the region; deformation energy measures the magnitude of the fitted change without cancellation between positive and negative deformations.
 Both \(p_{\mathcal R}^{PG}\) and \(p_{\mathcal R}^{DE}\) are calibrated by applying the same conditional-randomization null draws to their corresponding regional statistics. In structural benchmarks, we optionally use a gated local-rescue test
\begin{equation}
p_{\mathrm{global}}\le \alpha
    \quad\text{or}\quad
    \left(
    \alpha<p_{\mathrm{global}}\le \tau
    \ \text{and}\
    \min_{\mathcal R}p_{\mathcal R}<\alpha
    \right),
\end{equation}
where \(\tau>\alpha\) is a relaxed gate. The global conditional-randomization p-value remains the primary calibrated decision, while the local-rescue test targets weak structured alternatives near the global decision boundary. In the SDB experiments, \(\mathcal R\) is the oracle affected region induced by the known structural template, and \(p_{\mathcal R}\) is calibrated by the same conditional-randomization procedure on the regional statistic. Thus QRCIT+PG and QRCIT+DE evaluate regional testing with known affected regions. Proposition~2 does not establish calibration of the combined rescue rule. Learning \(\mathcal R\) and calibrating the resulting selection remain future work.

\noindent\textbf{Proposition 4 (Power under separated alternatives).}
Let
\begin{equation}
G_K=\frac{1}{K}\sum_{k=1}^K g(u_k)
\end{equation}
be the population predictive gain on \(\mathcal U_K\), and let \(c_{1-\alpha,K}\) be the oracle null critical value. Suppose \(T_K^{\mathrm{obs}}=G_K+o_p(1)\) and the conditional-randomization critical value consistently estimates \(c_{1-\alpha,K}\). If
\begin{equation}
G_K\ge c_{1-\alpha,K}+\gamma
\end{equation}
for some \(\gamma>0\), then
\begin{equation}
\Pr_{H_1}(p\le \alpha)\to 1.
\end{equation}

\noindent\emph{Sketch.}
By consistency, the observed statistic converges to \(G_K\), while the randomization critical value converges to \(c_{1-\alpha,K}\). Under the separation condition, the observed statistic eventually exceeds the null critical value with probability tending to one. The role of \(K\) is through \(G_K\): finer grids better approximate localized gain, although finite-sample power need not increase monotonically with \(K\).

\smallskip\noindent\textbf{Resolution efficiency and complexity.}
Resolution matters computationally only through evaluated quantile inputs. Since \(u\) is an input to \(\widehat Q_0\) and \(D_\eta\), increasing \(K\) does not increase the number of neural parameters. With a fixed architecture and sampled-\(u\) training, let \(F\) be the number of folds, \(E\) the number of epochs, \(B\) the number of null draws, \(n\) the sample size, and \(C_{\mathrm{bwd}},C_{\mathrm{fwd}}\) the costs of one backward and forward pass. Training scales as \(O(FEnC_{\mathrm{bwd}})\), while evaluation and calibration scale as \(O(F(B+1)nK C_{\mathrm{fwd}})\). Thus increasing quantile resolution incurs a linear forward-evaluation cost during calibration, but does not require a larger model or separate quantile heads.

\smallskip\noindent\textbf{Sliced multivariate extension.}
For multivariate \(X\), QRCIT applies the scalar test to one-dimensional projections
\begin{equation}
X^{(s)}=\theta_s^\top X,
    \qquad
    \theta_s\in\mathbb S^{d_x-1},
\end{equation}
and aggregates the resulting p-values across slices. This preserves the scalar quantile-flow machinery while extending QRCIT to multivariate responses.

\section{Experiments}
\label{sec:experiments}

\smallskip\noindent\textbf{Experimental goals.}
Our experiments evaluate four questions. First, we test whether QRCIT behaves as a calibrated conditional-independence test, with competitive empirical Type-I error and power on conventional synthetic post-nonlinear (PNL) benchmarks. Second, we evaluate whether QRCIT is sensitive to structured conditional-law changes through the proposed Structural Dependency Benchmark (SDB), where dependence may be localized in context, quantile region, sign, or distributional shape rather than appearing as a simple global mean shift. Third, we study whether quantile resolution and the local-rescue rule improve sensitivity to structured alternatives without introducing prohibitive computational cost. Fourth, we examine whether the learned quantile deformation provides useful real-world interpretation and downstream utility on Sachs, METABRIC breast cancer, and CCLE/PLX4720.

\smallskip\noindent\textbf{Setup.}
We compare QRCIT with representative kernel, regression, and neural CIT baselines, including GCM~\cite{shah2020gcm}, SGMCIT~\cite{ren2025sgmcit}, CDCIT~\cite{yang2024cdcit}, NNLSCIT~\cite{li2023nnlscit}, KCI~\cite{zhang2012kci}, and PowerMax-KCI~\cite{pogodin2024practical}. Synthetic results report empirical Type-I error under \(\beta=0\) and power under \(\beta=0.5\). Synthetic PNL and SDB comparisons use \(K=9\) quantile levels and \(B=20\) conditional-randomization null draws, so the rejection threshold is \(\alpha=1/(B+1)=1/21\approx0.05\). The SDB local-rescue variants QRCIT+PG/DE use the gate \(\tau=0.2\). Each PNL and merged-SDB point averages over 100 independent trials, while each separated SDB structural-pattern point averages over 20 trials per pattern. Sachs uses \(B=200\) and \(\alpha=1/201\approx0.005\), aligned with the finest attainable randomization p-value. METABRIC uses \(B=200\), giving p-value resolution \(1/201\), while CCLE uses \(B=1000\), with resolution \(1/1001\); in METABRIC, \(p<0.10\), \(p<0.05\), and \(p<0.01\) are feature-selection thresholds for downstream tumor-size regression rather than separate test-size settings. Real-world experiments use \(K=19\).

QRCIT uses 5-fold cross-fitting, predictive-gain statistics, and conditional-randomization calibration. For scalar continuous \(Y\), the standardized residual-bootstrap sampler fits a nuisance regression of \(Y\) on \(Z\), resamples standardized residuals, and reconstructs null draws \(\widetilde Y\mid Z\); multivariate continuous \(Y\) uses a learned Gaussian conditional sampler, and binary mutation variables use a logistic/Bernoulli sampler for \(\Pr(Y=1\mid Z)\). The neural quantile model uses two hidden layers of width 32, batch size 64, learning rate \(10^{-3}\), weight decay \(10^{-3}\), dropout \(0.05\), 80 training epochs, patience 20, \(\lambda_D=10^{-3}\), and \(\lambda_{\mathrm{mono}}=10^{-2}\). These settings were selected by a validation grid over hidden width, dropout, learning rate, and regularization weights. All experiments were run on compute nodes with 64GB CPU memory.

\smallskip\noindent\textbf{Synthetic PNL benchmarks and Structural Dependency Benchmark.}
We use the standard synthetic PNL benchmark as a conventional nonlinear CIT test bed before evaluating the proposed structural benchmark. Each trial is sampled from a pooled set of two nonlinear mechanism families. Both draw \(Z\sim\mathcal N(0,I_{d_Z})\), directions \(a_x,a_y\in\mathbb S^{d_Z-1}\), and independent noises \(\epsilon_x,\epsilon_y\sim\mathcal N(0,1)\). One family uses the latent construction
\begin{equation}
\begin{aligned}
\nu_y &= a_y^\top Z+\sigma_y\epsilon_y, \qquad
Y=g_y(\nu_y),\\
\nu_x &= a_x^\top Z+\beta\nu_y+\sigma_x\epsilon_x, \qquad
X=g_x(\nu_x),
\end{aligned}
\end{equation}
with \(g_x,g_y\in\{\mathrm{id},\tanh,\operatorname{asinh},\mathrm{cube}\}\). Because these transformations are monotone, this family has an analytic conditional-quantile structure and serves as a controlled benchmark for quantile-based testing; QRCIT learns the conditional quantile functions from the sampled data. The second family uses
\begin{equation}
\begin{aligned}
\eta_x &= a_x^\top Z+\sigma_x\epsilon_x,\qquad X=g_x(\eta_x),\\
\eta_y &= a_y^\top Z+\beta s(X)+\sigma_y\epsilon_y,\qquad Y=g_y(\eta_y),
\end{aligned}
\end{equation}
where \(s(X)\) is the standardized \(X\)-signal and \(g_x,g_y\in\{\mathrm{id},\mathrm{square},\mathrm{cube},\tanh,\cos\}\). This second family includes nonmonotone transformations and harder detection cases, so dependence is not reducible to a simple monotone quantile shift. The null uses \(\beta=0\), and the main alternative uses \(\beta=0.5\).

\begin{figure}[t]
\centering
\includegraphics[width=\linewidth]{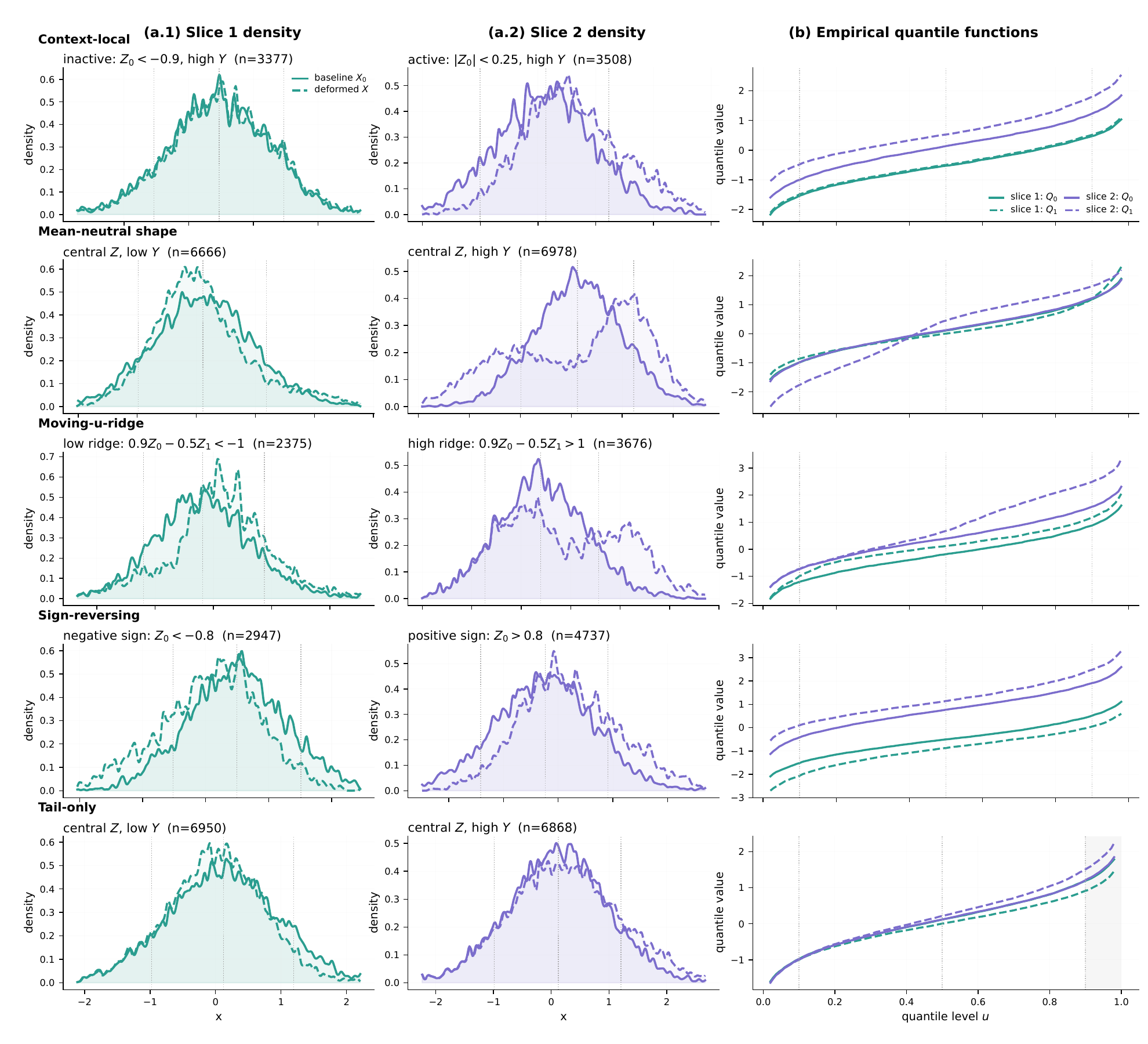}
\caption{
Structural dependence patterns in SDB. Each row shows one controlled structural alternative under the shared form \(X=Q_0(Z,U)+\beta d_s(Y,Z,U)\): context-local, mean-neutral shape, moving-\(u\)-ridge, sign-reversing, and tail-only dependence. The first two columns show density slices under different contexts, while the third column shows the corresponding quantile-function change. The benchmark stresses whether a CIT method can detect conditional-law changes that are localized over context \(Z\), localized over quantile level \(u\), sign-changing, or weak at the conditional mean.
}
\label{fig:sdb_gallery}
\end{figure}

We first visualize SDB in \figref{fig:sdb_gallery}. SDB is designed to isolate structural difficulty in CIT: signal strength may be comparable, but dependence can be localized in context, localized in quantile level, sign-changing, mean-neutral, or moving across the response distribution. SDB uses a common quantile-deformation generator,
\begin{equation}
Z\sim\mathcal N(0,I_{d_Z}),\quad
Y=m_Y(Z)+\epsilon_Y,\quad
U\sim{\rm Unif}(\varepsilon,1-\varepsilon),
\end{equation}
\begin{equation}
\begin{aligned}
X &= Q_0(Z,U)+\beta d_s(Y,Z,U),\\
Q_0(Z,U) &= \mu_0(Z)+\sigma_0(Z)\Phi^{-1}(U).
\end{aligned}
\end{equation}
Here \(s\) indexes the structural pattern and \(\beta=0\) gives the null. All alternatives share the same baseline conditional quantile law \(Q_0\); the pattern enters only through the structural deformation \(d_s\). Context-local dependence gates \(d_s\) to a slab in \(Z_0\), so global aggregation mixes active and inactive contexts. Sign-reversing dependence multiplies \(d_s\) by a sign-changing factor such as \(\tanh(1.25Z_0)\), so opposite regions can cancel. Tail-only dependence activates \(d_s\) only in high quantiles, so central quantiles appear nearly independent. Mean-neutral shape dependence uses positive and negative quantile deformation with approximately zero integral over \(u\), leaving the conditional mean weakly changed while changing the distributional shape. Moving-\(u\)-ridge dependence uses a context-dependent active quantile center \(u_\star(Z)=0.20+0.60\sigma(0.9Z_0-0.5Z_1)\), so the affected quantile band moves with \(Z\). These patterns serve as controlled probes of structural dependence: they test whether a method can retain power when dependence is concentrated in context, quantile level, sign, or distributional shape before being aggregated into a global statistic. SDB isolates these structural difficulties in quantile-deformation coordinates; PNL and real-data experiments assess performance beyond this controlled construction.

\begin{figure*}[t]
\centering
\includegraphics[width=\textwidth]{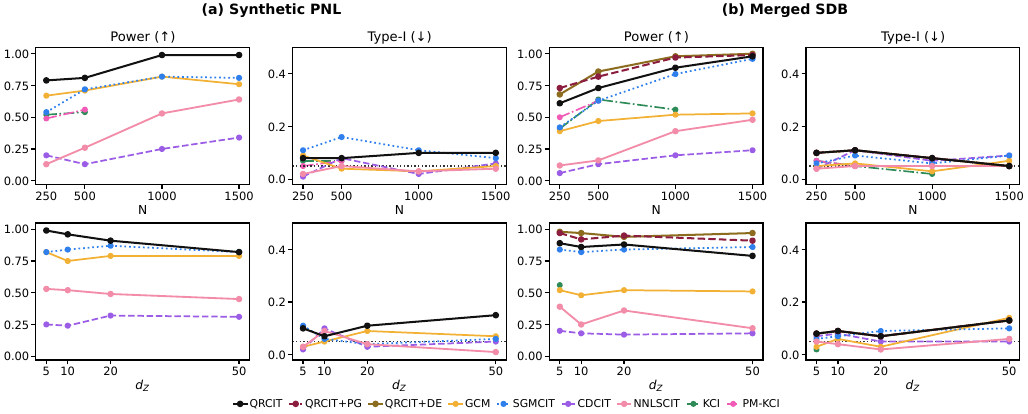}
\caption{
Type-I error and power on synthetic PNL and merged SDB benchmarks using \(K=9\), averaged over 100 independent trials. The first row varies sample size \(N\) with fixed \(d_Z=5\); the second row varies \(d_Z\) with fixed \(N=1000\). Panels (a) report synthetic PNL results and panels (b) report merged SDB results. QRCIT is highlighted; QRCIT+PG/DE use oracle regions to evaluate the benefit of regional testing.
}
\label{fig:pnl_sdb_main}
\end{figure*}

\begin{figure}[t]
\centering
\includegraphics[width=0.5\linewidth]{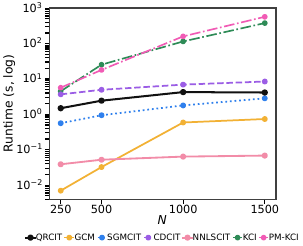}
\caption{
Average runtime on synthetic PNL as \(N\) increases with fixed \(d_Z=5\), shown on a log scale.
}
\label{fig:pnl_runtime}
\end{figure}

This design separates conventional Type-I control from power under structurally localized or cancelling dependence. \figref{fig:pnl_sdb_main} compares QRCIT with baselines on synthetic PNL and merged SDB. QRCIT remains competitive on PNL and strong on merged SDB. QRCIT+PG/DE quantify the benefit of using known affected regions in SDB. KCI and PowerMax-KCI are omitted in some larger settings because their runtime was prohibitive; their runtime trend is reported in \figref{fig:pnl_runtime}.

\begin{figure*}[t]
\centering
\includegraphics[width=\textwidth]{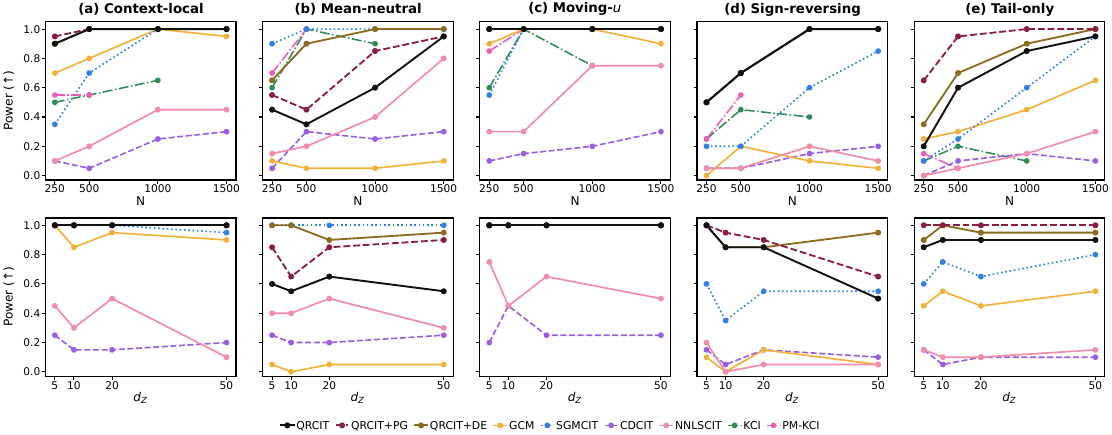}
\caption{
Separated SDB power, averaged over 20 trials per structural pattern. Each column corresponds to one structural pattern. The first row varies \(N\) with fixed \(d_Z=5\), and the second row varies \(d_Z\) with fixed \(N=1000\). The figure reports power only; Type-I behavior is summarized in \figref{fig:pnl_sdb_main}.
}
\label{fig:sdb_separated_power}
\end{figure*}

\figref{fig:sdb_separated_power} separates SDB performance by structural pattern. This view shows that the merged-SDB gains are not driven by a single easy case. QRCIT remains strong on context-local, moving-\(u\)-ridge, sign-reversing, and tail-only alternatives, while the mean-neutral case is more difficult and benefits from higher resolution and local rescue. Some baselines perform well on selected patterns but are less stable across the full structural benchmark.

\smallskip\noindent\textbf{Resolution, local rescue, and efficiency.}
\tabref{tab:local_rescue_pvalues} shows that testing within known affected regions yields smaller mean p-values under structured alternatives.

\begin{table}[t]
\centering
\caption{
Local-rescue p-value summary on merged SDB with fixed \(d_Z=5\) and \(K=9\). The null column reports mean global p-values under \(\beta=0\); the alternative columns report mean p-values under \(\beta=0.5\). Null calibration is assessed by rejection frequencies in \figref{fig:pnl_sdb_main}. Region p-values use oracle structural regions.
}
\label{tab:local_rescue_pvalues}
\resizebox{0.78\linewidth}{!}{
\begin{tabular}{c|c|ccc}
\toprule
& Null & \multicolumn{3}{c}{Alternative} \\
\cmidrule(lr){2-2}\cmidrule(lr){3-5}
\(N\)
& \(p_{\mathrm{global}}\)
& \(p_{\mathrm{global}}\downarrow\)
& \(p_{\mathcal R}^{PG}\downarrow\)
& \(p_{\mathcal R}^{DE}\downarrow\) \\
\midrule
250  & 0.650 & 0.250 & 0.179 & 0.160 \\
500  & 0.629 & 0.198 & 0.150 & 0.096 \\
1000 & 0.655 & 0.082 & 0.067 & 0.052 \\
\bottomrule
\end{tabular}
}
\end{table}

Finer quantile resolution improves power most clearly at moderate sample sizes (\tabref{tab:sdb_resolution_ablation}). Its benefit is small with few observations, while at larger sample sizes even the coarsest grid approaches maximal power. Thus the value of additional quantile detail depends on how much information the sample provides. \figref{fig:pnl_runtime} shows the computational tradeoff: QRCIT is more expensive than lightweight residual tests but remains practical compared with the heavier kernel methods.

\begin{table}[t]
\centering
\caption{
Quantile-resolution ablation on merged SDB with fixed \(d_Z=5\). Entries report global QRCIT power.
}
\label{tab:sdb_resolution_ablation}
\scriptsize
\setlength{\tabcolsep}{2.0pt}
\renewcommand{\arraystretch}{0.92}
\resizebox{0.55\linewidth}{!}{
\begin{tabular}{c|ccc}
\toprule
\(N\) & \(K=5\) & \(K=9\) & \(K=19\) \\
\midrule
250  & 0.60 & 0.61 & 0.60 \\
500  & 0.70 & 0.73 & 0.75 \\
1000 & 0.87 & 0.89 & 0.91 \\
1500 & 0.98 & 0.98 & 0.98 \\
\bottomrule
\end{tabular}
}
\end{table}

\smallskip\noindent\textbf{Additional ablations.}
We assess performance with multivariate responses and sensitivity to exchanging \(X\) and \(Y\). The sliced multivariate experiment uses \(n=1000\), \(d_X=d_Y=d_Z=5\), \(\beta=0.5\), 8 random plus coordinate projections, scalar QRCIT per slice with \(B=20\) and \(K=5\), a learned Gaussian \(Y\mid Z\) sampler, and Cauchy aggregation of slice-level p-values, without local-region tests.

\tabref{tab:sliced_multivariate} shows that one-dimensional projections allow QRCIT to detect dependence in multivariate responses. Exchanging \(X\) and \(Y\) leaves power nearly unchanged in the PNL experiment, but changes empirical Type-I error (\tabref{tab:role_swap}). The choice of which variable to model and which to resample can therefore affect calibration even when detection power is similar.

\begin{table}[t]
\centering
\caption{
Sliced multivariate benchmark. QRCIT applies the scalar quantile-flow test to one-dimensional projections of multivariate \(X\). Rows report empirical Type-I error and power.
}
\label{tab:sliced_multivariate}
\scriptsize
\setlength{\tabcolsep}{2.4pt}
\renewcommand{\arraystretch}{0.92}
\resizebox{0.9\linewidth}{!}{
\begin{tabular}{lccccccc}
\toprule
Metric & QRCIT & GCM & SGMCIT & CDCIT & KCI & PM-KCI & NNLSCIT \\
\midrule
Type-I \(\downarrow\) & 0.00 & 0.05 & 0.91 & 0.04 & 0.05 & 0.05 & N/A \\
Power \(\uparrow\)  & 1.00 & 1.00 & 1.00 & 0.24 & 1.00 & 1.00 & N/A \\
\bottomrule
\end{tabular}
}
\end{table}

\begin{table}[t]
\centering
\caption{
Role-swap ablation on PNL with \(N=1000\), \(d_Z=5\), and \(K=9\). Entries report empirical Type-I error and power before and after exchanging \(X\) and \(Y\).
}
\label{tab:role_swap}
\setlength{\tabcolsep}{4.5pt}
\renewcommand{\arraystretch}{0.95}
\resizebox{0.45\linewidth}{!}{
\begin{tabular}{lcc}
\toprule
Orientation & Type-I \(\downarrow\) & Power \(\uparrow\) \\
\midrule
Original & 0.10 & 0.99 \\
Swapped  & 0.04 & 0.98 \\
\bottomrule
\end{tabular}
}
\end{table}

\smallskip\noindent\textbf{Real-world benchmarks.}
We evaluate QRCIT on three real-world benchmarks with dataset-specific preprocessing.

\begin{itemize}
\setlength{\itemsep}{0ex}
\setlength{\parsep}{0pt}
\setlength{\parskip}{0pt}
\item \textbf{Sachs.} We use the Sachs protein-signaling benchmark from single-cell signaling measurements~\cite{sachs2005causal} and follow NNLSCIT preprocessing~\cite{li2023nnlscit}, yielding 90 CI/CD cases (50 CI, 40 CD). We report precision, recall, and F1 with conditional independence as the positive class; KCI is omitted due to the cost of repeated kernel tests.
\item \textbf{METABRIC breast cancer.} We reconstruct a CDCIT-style METABRIC benchmark from public cBioPortal \texttt{brca\_metabric} clinical files and mRNA z-scores~\cite{curtis2012metabric,yang2024cdcit}. Our reconstruction includes 1491 patients, 169 genes, and five clinical covariates; its cohort differs from the 1077 ER+ patients reported in CDCIT. QRCIT uses \(X=\) tumor size, \(Y=\) candidate gene, and \(Z=\) remaining genes plus covariates. The MSE scale is therefore not directly comparable to CDCIT; within this reconstruction, each method is evaluated with XGBoost grid-search CV. KCI is omitted because repeated kernel CI testing over the sample/feature loop is computationally prohibitive.
\item \textbf{CCLE/PLX4720.} We use the GCIT mutation-response benchmark~\cite{bellot2019gcit,barretina2012ccle}, with PLX4720 Amax as the response. Aligning mutation and response tables gives 474 cell lines. An absolute Pearson correlation screen at 0.1 retains 54 mutation features; adding the reported GCIT candidates and BRAF/MAPK indicators gives 75 features. QRCIT uses \(X=\) Amax, \(Y=\) candidate mutation status, \(Z=\) the other 74 mutations, and a logistic/Bernoulli \(Y\mid Z\) sampler. We report all ten candidates at seed 0 with five-fold cross-fitting, \(K=19\), and \(B=1000\). Because screening uses the response, these comparisons are exploratory; p-values are unadjusted for multiple testing.
\end{itemize}

On Sachs, QRCIT achieves the second-highest F1 among the compared methods (0.800), with recall of 0.880 (\tabref{tab:sachs}).

\begin{table}[t]
\centering
\caption{
Sachs real-data benchmark. Conditional independence is the positive class for precision, recall, and F1 over 90 CI/CD query cases. Blue indicates the best value in each row, and green indicates the second-best value.
}
\label{tab:sachs}
\scriptsize
\setlength{\tabcolsep}{3.2pt}
\renewcommand{\arraystretch}{0.95}
\resizebox{0.9\linewidth}{!}{
\begin{tabular}{lccccc}
\toprule
Metric & QRCIT & NNLSCIT & GCM & SGMCIT & CDCIT \\
\midrule
Precision \(\uparrow\) & 0.733 & \cellcolor{secondcell}0.746 & 0.676 & \cellcolor{bestcell}0.776 & 0.700 \\
Recall \(\uparrow\)    & 0.880 & \cellcolor{bestcell}0.940 & \cellcolor{secondcell}0.920 & 0.760 & 0.840 \\
F1 \(\uparrow\)        & \cellcolor{secondcell}0.800 & \cellcolor{bestcell}0.832 & 0.780 & 0.768 & 0.764 \\
\bottomrule
\end{tabular}
}
\end{table}

\tabref{tab:breast_cancer_mse} reports downstream tumor-size prediction on the reconstructed METABRIC benchmark. QRCIT-selected genes achieve the lowest mean MSE (\(0.965\pm0.317\)), with small differences among methods. \figref{fig:intro}A illustrates the added interpretation layer: SF3B1 shows an upward conditional tumor-size deformation, mildly stronger in the upper quantiles and consistent with splicing-factor dysregulation in breast cancer~\cite{tcga2012breast}; PTPRN2 shows a downward upper-tail deformation and is interpreted as a progression-associated conditional signature~\cite{tavazoie2008micrornas}.

\begin{table}[t]
\centering
\caption{
Downstream tumor-size prediction on the reconstructed METABRIC breast-cancer benchmark. Each cell reports mean MSE with standard deviation on the second line; lower is better. Q.10/Q.05/Q.01 denote QRCIT-selected genes at \(p<0.10\), \(p<0.05\), and \(p<0.01\). Blue indicates the best value, and green indicates the second-best value.
}
\label{tab:breast_cancer_mse}
\scriptsize
\setlength{\tabcolsep}{2.4pt}
\renewcommand{\arraystretch}{1.10}
\resizebox{0.9\linewidth}{!}{
\begin{tabular}{lccccccc}
\toprule
 & Q.10 & Q.05 & Q.01 & SGMCIT & CDCIT & NNLSCIT & GCM \\
\midrule
MSE \(\downarrow\)
&
\cellcolor{bestcell}\begin{tabular}{@{}c@{}}0.965\\[-0.2ex]\(\pm0.317\)\end{tabular}
&
\cellcolor{secondcell}\begin{tabular}{@{}c@{}}0.972\\[-0.2ex]\(\pm0.319\)\end{tabular}
&
\begin{tabular}{@{}c@{}}0.974\\[-0.2ex]\(\pm0.317\)\end{tabular}
&
\begin{tabular}{@{}c@{}}0.977\\[-0.2ex]\(\pm0.295\)\end{tabular}
&
\begin{tabular}{@{}c@{}}0.981\\[-0.2ex]\(\pm0.324\)\end{tabular}
&
\begin{tabular}{@{}c@{}}0.985\\[-0.2ex]\(\pm0.297\)\end{tabular}
&
\begin{tabular}{@{}c@{}}0.990\\[-0.2ex]\(\pm0.300\)\end{tabular}
\\
\bottomrule
\end{tabular}
}
\end{table}

On CCLE/PLX4720, QRCIT yields small p-values for BRAF.V600E, BRAF.MC, CDC42BPA, and PRKD1, supporting their conditional association with drug response (\tabref{tab:ccle_gcit}). Both QRCIT and GCIT yield large p-values for FLT3. QRCIT gives a more conservative assessment than GCIT in this application: for DNMT1, THBS3, and MAP3K5, it finds little evidence of additional response information after accounting for the other selected mutations.

The weak evidence for a MAP3K5 association is consistent with DGCIT~\cite{shi2021dgcit}. Biologically, MAP3K5/ASK1 is associated with stress-activated JNK/SAPK signaling, whereas PLX4720 targets the RAF--MEK--ERK pathway~\cite{ncbi_map3k5,bahar2023rafmapk}. Their involvement in different signaling branches offers context for the disagreement with GCIT. As noted in earlier analyses~\cite{bellot2019gcit,shi2021dgcit}, the true conditional relationships in this benchmark are unknown. Together with the limited cell-line sample, this makes the comparison exploratory.

\figref{fig:intro}B complements these results by describing the estimated response changes across quantiles. PIP5K1A\_MUT shows a positive fitted Amax-scale contrast across the quantile grid, whereas HIP1\_MUT shows a negative contrast that attenuates toward the upper quantiles. These genes have been linked to RAS-associated signaling and receptor trafficking, respectively~\cite{adhikari2018pip5k1a,rao2003hip1,bradley2007hip1egfr}. The profiles reveal the direction and distributional shape of the fitted response changes, providing an interpretation beyond the scalar p-values.

\begin{table}[t]
\centering
\caption{
CCLE/PLX4720 comparison on mutation indicators reported in GCIT. Entries are unadjusted p-values for conditional mutation--response associations. GCIT values are quoted from~\cite{bellot2019gcit} and use a different conditioning set.
}
\label{tab:ccle_gcit}
\scriptsize
\setlength{\tabcolsep}{2.4pt}
\renewcommand{\arraystretch}{0.95}
\begin{tabular*}{0.9\linewidth}{@{\extracolsep{\fill}}lccccc@{}}
\toprule
 & BRAF.V600E & BRAF.MC & HIP1 & CDC42BPA & THBS3 \\
\midrule
QRCIT \(p\) & 0.014 & 0.008 & 0.075 & 0.028 & 0.901 \\
GCIT \(p\)  & \(<0.001\) & \(<0.001\) & 0.008 & 0.050 & 0.013 \\
\bottomrule
\end{tabular*}

\vspace{0.8ex}

\begin{tabular*}{0.9\linewidth}{@{\extracolsep{\fill}}lccccc@{}}
\toprule
 & DNMT1 & PRKD1 & FLT3 & PIP5K1A & MAP3K5 \\
\midrule
QRCIT \(p\) & 0.956 & 0.013 & 0.912 & 0.118 & 1.000 \\
GCIT \(p\)  & 0.020 & 0.002 & 0.521 & 0.001 & \(<0.001\) \\
\bottomrule
\end{tabular*}
\end{table}

\section{Conclusion and Future Work}
\label{sec:conclusion}

We introduced QRCIT, a conditional independence test that describes how the distribution of \(X\) changes when \(Y\) is observed in addition to \(Z\). Its learned quantile deformation complements a global test by showing how these changes vary across contexts and quantile levels. We also introduced SDB, a controlled benchmark for evaluating sensitivity to different patterns of conditional dependence. Experiments show competitive performance on conventional benchmarks and improved detection of structured alternatives. The real-data studies illustrate the value of combining testing with distributional interpretation.

\smallskip\noindent\textbf{Relevance to Data Mining and Broad Impact.}
From a data-mining perspective, QRCIT targets a common setting in high-dimensional observational data: dependence is often not global, but concentrated in subpopulations, tails, or context-specific regimes. Such structure is important in biomedical discovery, treatment-response analysis, feature screening, and causal discovery, where a scalar rejection decision may be insufficient for understanding why a variable matters. By combining calibrated testing with quantile- and context-resolved deformation, QRCIT provides both statistical evidence and an inspectable description of where the conditional law changes. This makes the method useful not only as a hypothesis test, but also as a diagnostic tool for discovering weak, heterogeneous, or localized dependence patterns in complex data.

\smallskip\noindent\textbf{Future work.}
A key direction is to learn affected context--quantile regions from the estimated deformation field, motivated by the gains of oracle-region QRCIT over global QRCIT on SDB. Attention mechanisms or soft gating networks could use deformation features to assign weights across contexts and quantile levels, approximating the localization currently supplied by the oracle. Calibrating the resulting adaptive tests while accounting for region selection remains an important challenge~\cite{gablenz2026local}. Selecting and testing regions on the same data can favor patterns caused by random noise, so calibration must preserve the intended Type-I error rate despite this data-dependent selection. Further directions include adaptive quantile resolution and multivariate deformation models that capture changes in joint response structure.


\begingroup\footnotesize
\begin{thebibliography}{99}
\setlength{\itemsep}{0pt}\setlength{\parsep}{0pt}\setlength{\parskip}{0pt}

\bibitem[Giorgi et~al.(2014)]{giorgi2014cindy}
F. M. Giorgi, G. Lopez, J. H. Woo, B. Bisikirska, A. Califano, and M. Bansal, ``Inferring protein modulation from gene expression data using conditional mutual information,'' \emph{PLOS ONE}, vol. 9, no. 10, e109569, 2014.

\balance

\bibitem[Barretina et~al.(2012)]{barretina2012ccle}
J. Barretina et~al., ``The Cancer Cell Line Encyclopedia enables predictive modelling of anticancer drug sensitivity,'' \emph{Nature}, vol. 483, no. 7391, pp. 603--607, 2012.

\bibitem[Sachs et~al.(2005)]{sachs2005causal}
K. Sachs, O. Perez, D. Pe'er, D. A. Lauffenburger, and G. P. Nolan, ``Causal protein-signaling networks derived from multiparameter single-cell data,'' \emph{Science}, vol. 308, no. 5721, pp. 523--529, 2005.

\bibitem[Curtis et~al.(2012)]{curtis2012metabric}
C. Curtis et~al., ``The genomic and transcriptomic architecture of 2,000 breast tumours reveals novel subgroups,'' \emph{Nature}, vol. 486, pp. 346--352, 2012.

\bibitem[The Cancer Genome Atlas Network(2012)]{tcga2012breast}
The Cancer Genome Atlas Network, ``Comprehensive molecular portraits of human breast tumours,'' \emph{Nature}, vol. 490, pp. 61--70, 2012.

\bibitem[Tavazoie et~al.(2008)]{tavazoie2008micrornas}
S. F. Tavazoie, C. Alarc\'{o}n, T. Oskarsson, D. Padua, Q. Wang, P. D. Bos, W. L. Gerald, and J. Massagu\'{e}, ``Endogenous human microRNAs that suppress breast cancer metastasis,'' \emph{Nature}, vol. 451, pp. 147--152, 2008.

\bibitem[Adhikari and Counter(2018)]{adhikari2018pip5k1a}
H. Adhikari and C. M. Counter, ``Interrogating the protein interactomes of RAS isoforms identifies PIP5K1A as a KRAS-specific vulnerability,'' \emph{Nature Communications}, vol. 9, article 3646, 2018.

\bibitem[Rao et~al.(2003)]{rao2003hip1}
D. S. Rao et~al., ``Altered receptor trafficking in Huntingtin Interacting Protein 1-transformed cells,'' \emph{Cancer Cell}, vol. 3, no. 5, pp. 471--482, 2003.

\bibitem[Bradley et~al.(2007)]{bradley2007hip1egfr}
S. V. Bradley et~al., ``Huntingtin interacting protein 1 is a novel brain tumor marker that associates with epidermal growth factor receptor,'' \emph{Cancer Research}, vol. 67, no. 8, pp. 3609--3615, 2007.

\bibitem[NCBI Gene(2026)]{ncbi_map3k5}
NCBI Gene, ``MAP3K5 mitogen-activated protein kinase kinase kinase 5 [Homo sapiens],'' Gene ID: 4217, National Center for Biotechnology Information, accessed 2026-06-07.

\bibitem[Bahar et~al.(2023)]{bahar2023rafmapk}
M. E. Bahar, H.-J. Kim, and D.-R. Kim, ``Targeting the RAS/RAF/MAPK pathway for cancer therapy: from mechanism to clinical studies,'' \emph{Signal Transduction and Targeted Therapy}, vol. 8, article 455, 2023, doi: 10.1038/s41392-023-01705-z.

\bibitem[Adrian and Brunnermeier(2016)]{adrian2016covar}
T. Adrian and M. K. Brunnermeier, ``CoVaR,'' \emph{American Economic Review}, vol. 106, no. 7, pp. 1705--1741, 2016.

\bibitem[Adrian et~al.(2019)]{adrian2019vulnerable}
T. Adrian, N. Boyarchenko, and D. Giannone, ``Vulnerable growth,'' \emph{American Economic Review}, vol. 109, no. 4, pp. 1263--1289, 2019.

\bibitem[Chernozhukov and Hansen(2005)]{chernozhukov2005ivqr}
V. Chernozhukov and C. Hansen, ``An IV model of quantile treatment effects,'' \emph{Econometrica}, vol. 73, no. 1, pp. 245--261, 2005.

\bibitem[Bitler et~al.(2014)]{bitler2016headstart}
M. P. Bitler, H. W. Hoynes, and T. Domina, ``Experimental evidence on distributional effects of Head Start,'' NBER Working Paper 20434, 2014, doi: 10.3386/w20434.

\bibitem[Cand\`es et~al.(2018)]{candes2018knockoffs}
E. Cand\`es, Y. Fan, L. Janson, and J. Lv, ``Panning for gold: Model-X knockoffs for high-dimensional controlled variable selection,'' \emph{JRSS-B}, vol. 80, no. 3, pp. 551--577, 2018.

\bibitem[Hardt et~al.(2016)]{hardt2016equality}
M. Hardt, E. Price, and N. Srebro, ``Equality of opportunity in supervised learning,'' in \emph{Proc. NeurIPS}, 2016.

\bibitem[Peters et~al.(2016)]{peters2016invariant}
J. Peters, P. B\"uhlmann, and N. Meinshausen, ``Causal inference by using invariant prediction: identification and confidence intervals,'' \emph{Journal of the Royal Statistical Society: Series B}, vol. 78, no. 5, pp. 947--1012, 2016.

\bibitem[Zhang et~al.(2011)]{zhang2012kci}
K. Zhang, J. Peters, D. Janzing, and B. Schölkopf, ``Kernel-based conditional independence test and application in causal discovery,'' in \emph{Proc. UAI}, 2011, pp. 804--813.

\bibitem[Kim et~al.(2022)]{kim2021localperm}
I. Kim, M. Neykov, S. Balakrishnan, and L. Wasserman, ``Local permutation tests for conditional independence,'' \emph{The Annals of Statistics}, vol. 50, no. 6, pp. 3388--3414, 2022.

\bibitem[Hwang et~al.(2023)]{hwang2023local}
I. Hwang, Y. Kwak, Y.-J. Song, B.-T. Zhang, and S. Lee, ``On discovery of local independence over continuous variables via neural contextual decomposition,'' in \emph{Proc. Conference on Causal Learning and Reasoning}, PMLR, pp. 448--472, 2023.

\bibitem[Pogodin et~al.(2024)]{pogodin2024practical}
R. Pogodin, A. Schrab, Y. Li, D. J. Sutherland, and A. Gretton, ``Practical kernel tests of conditional independence,'' arXiv:2402.13196, 2024.

\bibitem[Neykov et~al.(2024)]{neykov2024wasserstein}
M. Neykov, L. Wasserman, I. Kim, and S. Balakrishnan, ``Nearly minimax optimal Wasserstein conditional independence testing,'' \emph{Information and Inference: A Journal of the IMA}, vol. 13, no. 4, iaae033, 2024.

\bibitem[Su and White(2012)]{su2012quantile}
L. Su and H. L. White, ``Conditional independence specification testing for dependent processes with local polynomial quantile regression,'' in \emph{Essays in Honor of Jerry Hausman}, Advances in Econometrics, vol. 29, pp. 355--434, 2012.

\bibitem[Petersen and Hansen(2021)]{petersen2021partial}
L. Petersen and N. R. Hansen, ``Testing conditional independence via quantile regression based partial copulas,'' \emph{Journal of Machine Learning Research}, vol. 22, no. 70, pp. 1--47, 2021.

\bibitem[Belloni et~al.(2025)]{belloni2025qgm}
A. Belloni, M. Chen, and V. Chernozhukov, ``Quantile graphical models: Prediction and conditional independence with applications to systemic risk,'' \emph{Journal of Econometrics}, vol. 252, article 106100, 2025.

\bibitem[Gablenz et~al.(2026)]{gablenz2026local}
P. Gablenz, M. Sesia, T. Sun, and C. Sabatti, ``Searching for local associations while controlling the false discovery rate,'' arXiv:2412.02182v3, 2026.

\bibitem[Shah and Peters(2020)]{shah2020gcm}
R. D. Shah and J. Peters, ``The hardness of conditional independence testing and the generalised covariance measure,'' \emph{Annals of Statistics}, vol. 48, no. 3, pp. 1514--1538, 2020.

\bibitem[Duong and Nguyen(2022)]{duong2022lcit}
B. Duong and T. Nguyen, ``Conditional independence testing via latent representation learning,'' in \emph{Proc. IEEE International Conference on Data Mining (ICDM)}, 2022, pp. 121--130, doi: 10.1109/ICDM54844.2022.00022.

\bibitem[Li et~al.(2023)]{li2023nnlscit}
S. Li, Y. Zhang, H. Zhu, C. Wang, H. Shu, Z. Chen, Z. Sun, and Y. Yang, ``K-nearest-neighbor local sampling based conditional independence testing,'' in \emph{Advances in Neural Information Processing Systems}, vol. 36, pp. 23321--23344, 2023.

\bibitem[Sen et~al.(2017)]{sen2017mpcit}
R. Sen, A. T. Suresh, K. Shanmugam, A. G. Dimakis, and S. Shakkottai, ``Model-powered conditional independence test,'' in \emph{Proc. NeurIPS}, 2017.

\bibitem[Bellot and van der Schaar(2019)]{bellot2019gcit}
A. Bellot and M. van der Schaar, ``Conditional independence testing using generative adversarial networks,'' in \emph{Proc. NeurIPS}, 2019.

\bibitem[Shi et~al.(2021)]{shi2021dgcit}
C. Shi, T. Xu, W. Bergsma, and L. Li, ``Double generative adversarial networks for conditional independence testing,'' \emph{Journal of Machine Learning Research}, vol. 22, no. 285, pp. 1--32, 2021.

\bibitem[Yang et~al.(2025)]{yang2024cdcit}
Y. Yang, S. Li, Y. Zhang, Z. Sun, H. Shu, Z. Chen, and R. Zhang, ``Conditional diffusion models based conditional independence testing,'' in \emph{Proceedings of the AAAI Conference on Artificial Intelligence}, vol. 39, no. 21, pp. 22020--22028, 2025, doi: 10.1609/aaai.v39i21.34354.

\bibitem[Ren et~al.(2025)]{ren2025sgmcit}
Y. Ren, C. Jin, Y. Xia, L. Ke, L. Huang, H. Xue, H. Zhang, J. Guan, and S. Zhou, ``Score-based generative modeling for conditional independence testing,'' in \emph{Proc. ACM SIGKDD Conference on Knowledge Discovery and Data Mining (KDD)}, 2025, pp. 2410--2419, doi: 10.1145/3711896.3737118.

\bibitem[He et~al.(2025a)]{he2025transport}
C. He, Y. Gao, L. Zhu, and J. Huang, ``Conditional independence test based on transport maps,'' arXiv:2504.09567, 2025.

\bibitem[Zhao et~al.(2026)]{zhao2026fmcit}
S. Zhao, Y. Yang, S. Li, and K. Fukumizu, ``Fast flow matching based conditional independence tests for causal discovery,'' arXiv preprint arXiv:2602.08315, 2026.

\bibitem[He et~al.(2026)]{he2025hardness}
Z. He, R. Pogodin, Y. Li, N. Deka, A. Gretton, and D. J. Sutherland, ``On the hardness of conditional independence testing in practice,'' \emph{Advances in Neural Information Processing Systems}, vol. 38, pp. 172300--172339, 2026.

\bibitem[Behrmann et~al.(2019)]{behrmann2019invertible}
J. Behrmann, W. Grathwohl, R. T. Q. Chen, D. Duvenaud, and J.-H. Jacobsen, ``Invertible residual networks,'' in \emph{Proc. ICML}, 2019, pp. 573--582.

\bibitem[Chen et~al.(2019)]{chen2019residual}
R. T. Q. Chen, J. Behrmann, D. Duvenaud, and J.-H. Jacobsen, ``Residual flows for invertible generative modeling,'' in \emph{Proc. NeurIPS}, 2019.

\end{thebibliography}
\endgroup
\end{document}